\documentclass{article}
\usepackage{ijcai21}

\usepackage{times}

\usepackage{soul}
\usepackage{url}
\usepackage[hidelinks]{hyperref}
\usepackage[utf8]{inputenc}
\usepackage[small]{caption}
\usepackage{graphicx}
\usepackage{amsmath}
\usepackage{booktabs}
\usepackage{multirow}
\usepackage{amssymb}

\newcommand{\acad}{author}
\newcommand{\Acad}{Author}
\newcommand{\Q}{\mathcal{Q}}
\newcommand{\A}{\mathcal{A}}
\newcommand{\academic}{a}
\newcommand{\QT}{\mathcal{Q_T}}
\newcommand{\Unigrams}{\mathcal{U}}
\newcommand{\Bigrams}{\mathcal{B}}
\newcommand{\QUnigrams}{\mathcal{U_Q}}
\newcommand{\QBigrams}{\mathcal{B_Q}}
\newcommand{\Profile}{\mathcal{P}_a}

\newcommand{\computing}{Computing} 

\usepackage{xcolor}
\newcommand{\OC}[1]{\textcolor{black}{#1}}

\newcommand{\FT}[1]{\textcolor{black}{#1}}

\usepackage{fancyhdr}
\title{An Explanatory Query-Based Framework for Exploring Academic Expertise}

\author{
Oana Cocarascu$^1$\and
Andrew McLean$^2$\and
Paul French$^2$\And
Francesca Toni$^2$\\
\affiliations
$^1$King's College London\\
$^2$Imperial College London\\
\emails
oana.cocarascu@kcl.ac.uk,
\{andrew.mclean,paul.french,f.toni\}@imperial.ac.uk
}

\begin{document}

\maketitle

\begin{abstract}
The success of research institutions heavily relies upon identifying the right researchers ``for the job'': researchers may need to identify appropriate collaborators, often from across disciplines; students may need to identify suitable supervisors for projects of their interest; administrators may need to match funding opportunities with relevant researchers, and so on. Usually, finding potential collaborators in institutions is a time-consuming manual search task prone to bias. In this paper, we propose a novel query-based framework for searching, scoring, and exploring research expertise automatically, based upon processing abstracts of academic publications. Given user queries in natural language, our framework finds researchers with relevant expertise, making use of domain-specific knowledge bases and word embeddings. It also generates explanations for its recommendations. We evaluate
\FT{our framework with an institutional repository of papers from a leading university, using, as baselines, artificial neural networks and transformer-based models for a multilabel classification task to identify authors of publication abstracts. We also assess  } the cross-domain effectiveness of our framework with a (separate) research funding repository \FT{for the same institution}. We show that our simple method is effective in identifying matches, while satisfying desirable properties and being efficient.
\end{abstract}

\section{Introduction}

The success of research institutions heavily relies upon identifying the right researchers ``for the job''. For example, researchers may need to identify appropriate collaborators, 
as scientific, technological and other research often requires a diverse set of skills, frequently cross-disciplinary. The capability of identifying suitable research partnerships is crucial to academic and industrial researchers alike, independently of their level of seniority. Indeed, often, companies target researchers in universities to collaborate with; also, researchers and administrators may need to identify  potential cross-disciplinary collaborations within their institutions, and research administrators may need to match funding opportunities (from industry, private donors and/or funding bodies) with relevant researchers. As a further example, (current or prospective) students of academic institutions may need to identify suitable supervisors for projects of their interest. These tasks are usually performed manually: one reviews the research interests of researchers and their publications and then contacts the relevant \acad s, with academic search engines having become the go-to resource for finding research papers. Amongst the most popular platforms, we find dblp\footnote{\url{https://dblp.org/}}, Google Scholar\footnote{\url{https://scholar.google.com/}}, Microsoft Academic\footnote{\url{https://academic.microsoft.com/home}}, and Semantic Scholar\footnote{\url{ https://www.semanticscholar.org/}}.
But, as the global scientific output doubles every nine years\footnote{\url{https://blogs.nature.com/news/2014/05/global-scientific-output-doubles-every-nine-years.html}}, there is a need for this task to be automated. The automation may also help avoid biases towards researchers one is acquainted with already.

Various automated expertise retrieval systems have been proposed, useful in a range of settings such as expert profiling, expert finding, and expert recommendation \cite{Goncalves:19}. Nonetheless, several open issues remain \cite{Goncalves:19}, including: \textit{1)} the need of explanation of the results, in particular about how the relevance of researchers is assessed, and \textit{2)} the identification of matches in support of cross-domain collaborations. In this paper, we address both these issues. The ability to support cross-domain  collaborations leads not only to unique combinations of ideas but also allows research questions that have been addressed in one domain to be transferred to other domains, for example from computer vision to medicine.

We propose a novel framework for finding researchers with particular expertise to match users' queries in natural language, of any length, returning explanations for why the researchers were identified and supporting retrieval across domains. Within our framework, a user may search for specific, narrow forms of expertise, e.g., \emph{3D medical imaging segmentation}, but also for more general topics such as \emph{artificial intelligence}.
Given any user query as input, our framework identifies and assigns a score to researchers according to their relatedness to the query. It also returns explanations for its outputs, thereby helping users select researchers or refine queries. This is a feature that other works in the literature severely lack. Our explanations are in the form of unigrams and bigrams describing the match between the query and authors returned. Additionally, our method is not limited to a single domain, but is cross-disciplinary and can identify (joint) works, \FT{e.g. across} subjects such as Computer Science, Bioengineering, Medicine, Physics, Mathematics, and so on. 
Our method supports a retrieval, rather than a ranking task, as the latter may disadvantage early-career researchers (who typically have a less substantial publication record than senior researchers). Also, whilst oftentimes an expert can be deemed to be the author with the highest number of publications in the field, practical constraints are crucial for the task of finding researchers with relevant expertise. For example, governments or senior researchers would choose a professor, however, early-career researchers are more likely to benefit from collaborating with junior researchers. Our framework can accelerate a large variety of matchings, alongside the deployment of filters when career-stage (and other factors) should play a role in answering queries. 

Our framework \FT{relies upon a} repository of academic papers authored by the target researchers.  Thus, we use \emph{author} and \emph{researcher} interchangeably in the paper. The framework combines multiple measurements of expertise, derived from \textit{i)} textual content in the form of abstracts of scientific publications, \textit{ii)} word embeddings learnt from these abstracts, and \textit{iii)} domain-specific knowledge bases.
Our framework is simple but effective at identifying matches, while also satisfying desirable properties and being efficiently deployable. Specifically, we evaluate it
\FT{with data drawn from the internal  repository of a leading  research institution which focuses on STEM\footnote{Science, technology, engineering, and mathematics.} topics (namely Imperial College London\footnote{\url{http://www.imperial.ac.uk/}}), } against artificial neural networks and transformer-based models, as natural alternatives within the state-of-the-art in natural language processing. 
To do so we train feed-forward artificial neural networks (as baseline) and various state-of-the-art transformer-based models (variants of BERT \cite{bert}) for the (highly) multilabel task of predicting authors of publication abstracts.   Concretely, we experiment with:  BERT \cite{bert}, RoBERTa \cite{roberta}, ALBERT \cite{albert}, DistilBERT \cite{distilbert}, and SciBERT \cite{scibert}. We focus our evaluation on four different domains: Computer Science, Electronic/Electric Engineering, Bioengineering, Mathematics. For evaluation against all alternative models considered, the domains are considered separately in order to reduce the number of labels to a manageable level (from around 6000 labels in total to a maximum of 105 labels). 
For the  cross-domain evaluation, we use
a separate institutional research funding repository \FT{(for the same institution)}
and show  that our framework achieves high performance when tested with  titles of awarded grants (within and across disciplines) treated as queries. We show that, while BERT-base language models are performing reasonably well within domains, our novel framework performs well within and \emph{across} domains. Finally, we 
\FT{show, for the instance  obtained on the specific institutional repository used for the evaluation, that our framework satisfies (desirable)} 
properties of \emph{diversity} and \emph{novelty} in the matches it identifies for queries.

The remainder of the paper is organized as follows. We give an overview on related work in Section \ref{relatedwork}. We describe our framework for identifying academic expertise in Section~\ref{search}. We conduct extensive experimentation to identify academic expertise, using our proposed framework and artificial neural networks/transformer-based models, and report the results in Section \ref{evaluation}.
In Section~\ref{sec:new} we illustrate deployments of our framework and discuss efficiency  thereof. We draw conclusions and suggest directions for future work in Section \ref{conclusion}.
 
\section{Related work} \label{relatedwork}

The task of finding  experts can be broadly split into: creating profiles, identifying experts (e.g. using probabilistic or graph-based models), and matching queries to help find ``who knows what" \cite{Linsurvey:17}.
Several efforts have been made to tackle this
task (see \cite{Balog:12} for a survey). Some works focused on identifying experts within a university \cite{Balog:07,Cifariello:19} or a country \cite{Mangaravite:16}.
The method of \cite{Balog:07} modelled the probability of a candidate being an expert given the query topic and the probability of a knowledge area being part of the candidate's profile. The candidates can select their areas of expertise from a list of topics. Our framework on the other hand automatically learns the features that characterise candidates.
In \cite{Cifariello:19}, the authors created a labelled and weighted graph for each academic author where nodes are Wikipedia entities that occur in the academic's publications. They also used embeddings, which were learnt from Wikipedia. The authors of \cite{JunfengLiu:20} developed a scholarly search system that, given a set of keywords, returns a sorted list of articles whose titles contain the given keywords with relevance determined using pre-trained word embeddings \cite{Mikolov:13}. In contrast, our framework learns word embeddings from the publication abstracts, making the word vectors better match the \acad s.

The approach of \cite{Patel:20} generated candidate phrases of unigrams, bigrams, and trigrams consisting of nouns and adjectives from 30000 randomly selected ACM papers. In addition to a cross-validation evaluation, the authors also compared the extracted keyphrases of papers with the ones supplied by the papers' authors. The focus of \cite{Berger:20} was on analysing Computer Science papers for academic expert retrieval. Their method relies on Sentence-BERT \cite{Reimers:19} which is used on the concatenation of the title and the abstract to represent the paper. The citation counts of the papers was used as a proxy for expertise. Our dataset is not limited to Computer Science papers as we explore several domains and perform a cross-domain evaluation of our framework.

The authors of \cite{Tang:08} proposed a unified topic model for simultaneously modelling the topical distribution of papers, authors, and conferences. \cite{Gysel:16} modelled the conditional probability of a candidate's expertise given a single query term and used distributed vector representations to implement semantic matching between query terms and candidate profiles. Some works focused on identifying representative features such as \cite{Moreira:15}, who proposed different feature types for determining author expertise: profile information features (e.g. number of papers per year), textual similarity features, and features based on citation patterns. Others focused on algorithms using ranking \cite{Brochier:20} or graph-based models including collaborators, papers, citations, conferences \cite{Robertie:17,Ammar:18}. In this paper, we use textual data only (i.e. abstracts) to create \acad s profiles.

There are other works that have analysed publications but have addressed different research problems than in our paper. 
\cite{YiZhang:19} learnt embeddings for academic papers and used them for three tasks: paper classification of research fields, paper similarity, and paper influence prediction. \cite{Mai:18} proposed a retrieval system that returns the most similar papers to a given query and evaluated the system in two settings: paper similarity and co-author inference. Several works aimed to identify reviewers to be assigned to conference paper submissions which also involved creating an academic profile, similarly to our method for identifying academic expertise. Various approaches have been used, e.g. topic modelling \cite{Mimno:07,Charlin:13} and probability distribution over vocabulary \cite{Balog:06}. \cite{Charlin:13} analysed Computer Science publications and represented papers as word count vectors and reviewers as the normalized word count for each word appearing in the reviewers' papers. Also in the Computer Science domain, \cite{Anjum:19} used abstract topic vectors and word embeddings \cite{Mikolov:13} to derive semantic representations of paper abstracts and reviewers' past abstracts, with cosine similarity to measure the semantic similarity. In this paper, we do not limit our analysis to Computer Science but explore several domains and test the performance of our framework cross-domain. 

\cite{Silva:20} proposed an algorithm that corrects ``friendly citations" to improve author ranking. Several works have focused on paper recommender systems \cite{Kanakia:19,Waheed:19}. \cite{Waheed:19} used citation networks and co-authorship for recommending research papers. \cite{Kanakia:19} used co-citation and word embeddings learnt from publication abstracts for developing a hybrid paper recommender system which was evaluated with a user study. \cite{Shao:20} used graph convolutional networks to recommend relevant repositories on GitHub that match a given paper.

\begin{figure}[t]
	\centering
	\includegraphics[width=0.47\textwidth]{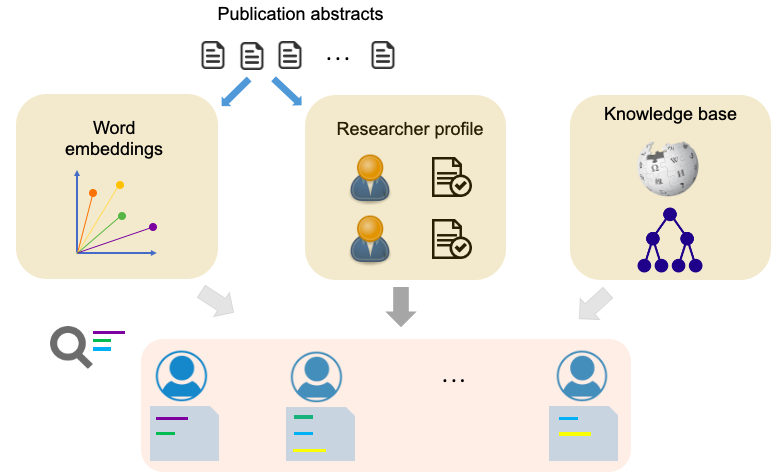}
	\caption{An overview of our framework. Abstracts of publications are used to learn word embeddings representative of our author dataset and to create profiles for researchers. We use Wikipedia glossaries for university subjects (e.g. \OC{\computing}) to construct a two-level knowledge base. Given a search query, our framework returns a list of \acad s obtained using  profiles. If specified, our framework may also use terms from the embeddings and/or knowledge base to identify researchers matching the query. For each identified researcher, our framework returns an \emph{explanation vector} describing how the researcher matches the query.}
	\label{fig:epic}
\end{figure}

Few works have addressed the task of cross-domain collaboration. \cite{JieTang:12} proposed a cross-domain topic learning model and evaluated on four cross-domain test cases: Data Mining to Theory, Medical Informatics to Database, Medical Informatics to Data Mining, and Visualization to Data Mining.
\cite{Scells:20} proposed a framework for automatically formulating Boolean queries for systematic review literature search. In this paper, we focus on processing publication abstracts and learn word embeddings from the dataset. Compared to previous works, we evaluate against a repository of funding granted to \acad s which is often cross-disciplinary. Additionally, we compare our framework against various state-of-the-art transformer-based models.

At their core, most existing works have attempted creating profiles that are then matched with queries.  This is also the case for the methodology in our proposed framework, while the specifics of our profile creation is bespoke, as is the definition and generation of explanations for recommendations.

\section{Framework for identifying academic expertise
} \label{search}

In this section we describe \FT{our framework, including the type of data it uses (in the form of \acad s' publication abstracts)}, the steps required for identifying academic expertise (amounting to creating academic profiles \FT{from the data}, using domain-specific knowledge bases and/or learning word embeddings to enhance performance) and our methodology for searching and exploring \acad s by their expertise given user queries in natural language. Figure \ref{fig:epic} shows an overview of our framework.

\subsection{\Acad\ data}
\label{sec:author data}

\FT{Our framework requires, as input data,} information about academics (\acad s) in \FT{the institution(s) of interest} and their publications. We make use of the following fields for \acad s: first name, last name, post, department, and faculty. For publications, we use: abstract, authors, and publication date. Note that we \FT{choose not to use} the titles of the publications as the information  therein is already included  in the abstracts and their removal did not affect performance \FT{in our experimental evaluation (see Section~\ref{evaluation})}, as \FT{our framework checks} whether n-grams occur, not \FT{their} frequencies \FT{(as detailed later in Section~\ref{sec:met})}. 


\subsection{Academic profile}

Our search method relies on extracting representative features to create a profile for each \acad. For this, we analyse the \acad's publication abstracts and consider  \acad s to be represented by the abstracts from their  publications.

Each abstract is represented as a text document. We first process the texts by lemmatizing the words and removing those with non-informative part of speech tags such as punctuation, numerals, determiners, and pronouns. We then convert the text documents to a matrix where each row indicates the presence or absence of a term (unigram or bigram) from the vocabulary extracted from the \acad's abstracts. We keep as features those terms that appear in at least two documents. For each feature, we keep track of the years of the publications in which they appear. Note that we are interested in the publication years as this information will be used to calculate the \acad s' \textit{academic score}, giving more importance to (terms in) more recent publications. This score may provide useful information when searching for \acad s, as we discuss later in Section~\ref{sec:met}. We apply this process to each \acad\ in our database to obtain the authors' profiles, and also index the features for faster information retrieval.

\begin{table*}
\centering
\begin{tabular}{ ccc } 
\textbf{Word} & \textbf{Similar words in word2vec} & \textbf{Similar words trained on the \acad\ dataset} \\ \hline \hline
\multirow{2}{*}{language} & languages, word, culture, english, & syntax, semantic, javascript, \\
& pronunciation, literature, programming & grammar, symbolic, syntactic, sentence \\ \hline 
\multirow{2}{*}{learn} & understand, learned, how, & neural, supervise, adversarial,  \\
& teach, tell, explain, things & classifier, cnn, ilp, encoder \\ \hline 
\end{tabular}
\caption{Differences in similar words between the original word2vec embeddings and our embeddings, in the context of our evaluation (see Section~\ref{evaluation}).}
\label{table:word2vec}
\end{table*}

\subsection{Enhancing the academic profile} \label{sec:extra}

We explore two ways to enhance the academic profiles: by incorporating external knowledge and by learning word embeddings from the profiles. We describe these next, and experiment with them (in isolation and in combination) in Section~\ref{evaluation}.

\subsubsection{Knowledge base}

\FT{Our framework allows for the integration of a knowledge base with taxonomic information about subjects of relevance to the institution(s) of interest. This knowledge base amounts to a graph }
 where the parent of a node in the graph is a more general term than any of its children.
For example, one node in the knowledge base \FT{may be}  ``computer science" \FT{including} amongst its children the following terms: ``artificial intelligence", ``machine learning", ``formal verification". Further, the node ``artificial intelligence" \FT{may include}, amongst others, the following \FT{children}: ``natural language processing", ``natural language generation", ``symbolic artificial intelligence", ``reinforcement learning", ``machine learning". 
As before, \FT{we envisage that} the terms
\FT{in the knowledge base are lemmatized} so that, for example, `framework' and `frameworks' \FT{are successfully matched}.

\subsubsection{Embeddings}
\label{sec:emb}

We use word2vec \cite{Mikolov:13} to learn word vectors representative of our author dataset so that similar words are close to each other in the vector space. Thus, the aim of the learnt vectors is to capture domain-specific lexical semantics. We first process the texts by lemmatizing the words and removing those with non-informative part of speech tags such as punctuation, numerals, determiners, and pronouns. To learn the embeddings, we use the Continuous Bag of Words (CBOW) model \cite{Mikolov:13}, where context words are used to predict a target word. \FT{In the evaluation in Section~\ref{evaluation} w}e use a window size of 5, ignore words with a frequency lower than 10, and learn vectors of dimension 100. These parameters were chosen experimentally for best performances.

We \FT{then} use the learnt embeddings to find similar words\footnote{We use `word' and `term' interchangeably in the paper.} for any given word, computed using cosine similarity between the word vector and the vectors for each word in the learnt embeddings. Table \ref{table:word2vec} shows some differences between original word2vec embeddings and learnt embeddings \FT{for the experiments in Section~\ref{evaluation}}. Note that \FT{here} the learnt embeddings \FT{reflect the STEM-based nature of our evaluation}. In particular, \FT{the learnt} embeddings  \FT{convey} similarities between `learn' (i.e. from ``machine learning") and `neural', `classifier', `cnn' (i.e. convolutional neural networks) rather than generally associated words that are found in word2vec such as `teach', `explain', `understand'. \OC{Word embeddings are useful to discover words that are used in similar contexts and thus can improve the task of finding academic expertise given a search query.}

\subsection{Methodology}
\label{method}
\label{sec:met}

Given a user query $\Q$ as input, our method returns a set of \acad s relevant to the query. In this section, we will use the following notations: $\A$ is the set comprising all \acad s,  $\Unigrams$ is the set of all unigrams that appear in \acad s' profiles, $\Bigrams$ is the set of all bigrams that appear in \acad s' profiles, and $\Profile$ gives the features (i.e. unigrams and bigrams) that represent \acad\ $a$'s profile. Note that we see the problem of identifying users matching the user query as a retrieval task and not as a ranking task, as the latter may lead to a bias against early career \acad s, who may be more likely, in some cases, to engage with the activities motivating the user queries than their over-committed senior counterparts with more experience who may be ranked higher. 

\subsubsection{Search}
We first lemmatize $\Q$ and extract all possible unigrams and bigrams. We then form the query search terms ($\QT$) by keeping only those unigrams and bigrams that are valid, namely members of $\Unigrams\cup \Bigrams$. Thus, $\QT =  \QUnigrams \cup \QBigrams $, where $ \QUnigrams \subseteq \Unigrams \bigwedge \QBigrams \subseteq \Bigrams$. The query terms in $\QT$ will be used to search for \acad s, against the academic profiles. The set of \acad s relevant to $\Q$ is then the set of all \acad s whose profiles include terms in $\QT$, as follows: 
\begin{center}
$R_{\Q} = \{\academic \in \A   | t \in \Profile \bigwedge   t \in \QT\}$. 
\end{center}
We score \acad s in $R_{\Q}$ for relevance to the query $\Q$ using both \textit{explanation and academic scores}, defined next for any \acad\ $\academic\in R_{\Q}$ using $\QT$. These scores may be useful to some users to explore top-ranked \acad s first, while providing information useful to explain why \acad s are returned as matches to queries.
\paragraph{\textbf{Explanation score.}}
We return an \emph{explanation vector} $V$ of features (unigrams and/or bigrams) that describes the match between $\QT$ and $a$. Then, we use $V$ to compute the explanation score, $S_E$, for $\academic$ where a bigram in $V$ gets a score of 10 and a unigram in $V$  gets a score of 1. The choice of these unit scores is driven by having a difference of an order of magnitude between bigrams and the unigrams that compose them as some words may appear in many, varied contexts (e.g. `analysis' and `mining'). Thus, for example, \emph{sentiment analysis} is given a higher score than either \emph{sentiment} or \emph{analysis}, leading to a better selection of relevant \acad s.
\paragraph{\textbf{Academic score.}}
We also compute an academic score, $S_A$,  for $\academic$, based on the year of publications of the papers from which the abstracts are drawn. Any publication that contains a term in $\QT$ and is older than 20 years receives a score of 0.01. Every year in the most recent 20 years (the current year included) receives a score based on its position in a row vector of 20 evenly spaced points between 0.05 and 1. For example, a paper from the year 2020 
receives a score of 1, whereas a paper from 2017 receives a score of 0.85. Note that the number of years considered (i.e. 20) for being assigned a score from evenly spaced points in an interval was set in order to accommodate various breakthroughs in the domains analysed but other values could be used.

\subsubsection{Explore}
We also explore other \acad s, possibly not in $R_{\Q}$, using our knowledge base and learnt word embeddings as described in Section \ref{sec:extra} to discover new terms to be added to $\QT$. We obtain a revised set of relevant \acad s $R_{\Q}^+$ (from  $R_\Q$).
\paragraph{\textbf{Knowledge base.}}
We extend $\QT$ to include any terms related to terms in $\QT$ that are found in the knowledge base. The result is a new set of terms $\QT^+$. We then repeat the search with each new term in turn and add the top 50 \acad s, based on the explanation and academic scores, to $R_\Q$ to obtain $R_{\Q}^+$. We found 50 to be a good number in terms of (new) authors returned.  Of course, the explanation score can be used to determine their relevance with respect to the query and help users filter out any that are not of interest.
\paragraph{\textbf{Embeddings.}}
We also replace each term in $\QT$ with its 25 most similar words as given by the learnt word vectors. We then repeat the search with the term replaced by the 25 similar words in turn.\footnote{This number was determined while inspecting similar words in the vector space.} Here as well we add to $R_\Q$ (or its modification to take knowledge into account, depending on whether we use also the knowledge base) the top 50 \acad s based on their scores. Note that, while we repeat the search several times for each new term obtained from the knowledge base or using embedding similarity, the search is efficient as we use PostgreSQL and indexing for storing the information used.
\paragraph{\textbf{Combining multiple occurrences of \acad s.}}
Note that $|R_{\Q}^+ \setminus R_\Q| \leq 50$, i.e. some of the 50 \acad s retrieved using the knowledge base and/or embeddings may already be in $R_\Q$. Moreover, \acad s that are identified using the knowledge base and/or the embeddings and that occur multiple times are only returned once, with their scores merged to create a single result. In particular, we concatenate the explanation vectors and sum up the explanation and the academic scores to obtain the new, merged scores.

\section{Evaluation} \label{evaluation}

In this section, we perform a number of experiments for our task of identifying academic expertise relevant to user queries. 
\FT{Our experiments use, as input data for our framework, an \acad\ dataset drawn form an institutional repository (see Section~\ref{sec:author data-i})
and a knowledge base drawn from Wikipedia (see Section~\ref{sec:kb-i}).}
In the experiments, we also use a \emph{grant dataset} (see Section~\ref{sec:data}) drawn from a separate institutional repository of funding granted to the university's \acad s, from a variety of national and international funding bodies and industries. Specifically, we use titles (with keywords concatenated, where present) of grants in this repository as user queries and aim to match them with the academics (\acad s) who were actually granted them. We conduct two sets of experiments. In the first experiments, we focus on four domains/departments within the university and compare our framework from Section~\ref{search} against feed-forward artificial neural networks and transformer-based models for author prediction within each department (Section~\ref{exp2}). 
In the second set of experiments,
we apply (variants of) our search method (with and without knowledge bases and embeddings) on the full grant dataset, across domains (Section~\ref{exp1}). Finally, we  explore some properties of our search methodology (Section~\ref{sec:properties}).

\subsection{\FT{\Acad\ dataset}}
\label{sec:author data-i}

\FT{We instantiate our framework with data from a high-ranked university in the world, \OC{namely Imperial College London}, which focuses on STEM subjects.\footnote{The data was accessed through an internal, private repository but could be reproduced from the publicly available list of academics in the university, using public repositories of publications.}
Our resulting dataset contains information about academics (\acad s) in this university and their publications, obtained as indicated in Section~\ref{sec:author data}.
Note that, for this institution, posts amount to professor, lecturer/assistant professor (at different levels of seniority), and research assistant/associate.
In summary, our \acad\ dataset contains information about 6,213 \acad s and 238,362 publications' abstracts. We use (fragments of) this dataset for training (our and other) models.}

\FT{Note that, even if we draw our data from a single institution, our \acad\ dataset provides a rich ground for academic search, as domain experts may be in competition (e.g. multiple labs may be working on natural language processing). Consequently, our framework is not limited to searching within a single cluster of academics working together, but rather looks at multiple groups, if they exist.}

\subsection{\FT{Knowledge base}}
\label{sec:kb-i}

\FT{We construct a knowledge base by combining the domain-specific knowledge bases obtained by querying the Wikipedia glossary associated with the specific subject of each existing department in the university.\footnote{For example, for the \OC{\computing} department, we use the glossary at \url{https://en.wikipedia.org/wiki/Glossary_of_computer_science}.} Further, we query the Wikipedia glossary with each term obtained, to generate a two-level-deep taxonomy per subject/department. }

\subsection{Grant dataset} \label{sec:data}

We conduct extensive evaluation using data about grants awarded to the university's \acad s, for which we gathered the title and, if present, the keywords.\footnote{From now on, even when not explicitly stated,  grant titles will be assumed to include these keywords (concatenated to the titles), if present.} We filter out grants whose titles have fewer than 5 words, that contain specific terms such as `studentship', 'bursary', `salary', and grants that contain date, budget amount, or names of organisations or persons, determined using named entity recognition. For example, we filter out grants such as ``PhD Studentship at \dots in the Area of \dots", ``Bursary supplement for \dots", ``Salary support for Professor \dots". This gives a total number of  2697 grants to be used for evaluation in both sets of  experiments.

\subsection{In-domain identification of \acad s} \label{exp2}

Here we compare the performance of our framework against feed-forward artificial neural networks (ANN) and transformer-based models in the task of \acad\ prediction, using the abstracts in the \acad\ dataset as training data. Given that, with more than 6000 authors in this dataset (see Section~\ref{sec:author data}),  author prediction as a multilabel classification task is not feasible, we  focus our comparison on authors from only four chosen domains (out of 15 domains).  

We select four domains/departments that overlap in terms of research: Computer Science (CS), Electronic/Electric  Engineering (EE), Bioengineering (Bioeng), and Mathematics (Maths), and use \acad s, abstracts and grants from these domains for the experiments in this section. The focus on restricted domains allows us to  reduce the number of \acad s from over 6000 in the full \acad\ dataset to a more manageable number of labels  for a classification task (see Table~\ref{table:stats}). This shows the number of abstracts for each of the four domains considered, with each abstract appearing as many times as the number of authors of the corresponding paper: thus, each abstract corresponds to a paper-\acad\ pair. Note that there may be multiple authors per paper, hence the number of papers is smaller than the number of paper-author pairs.

\begin{table*}
\centering
\begin{tabular}{ cccccc } 
\textbf{Domain} & \textbf{Abstracts} & \textbf{Papers} & \textbf{\Acad s} & \textbf{Mean papers/\acad} & \textbf{Median papers/\acad} \\ 
& \textbf{(paper-\acad\ pairs)} &  &  & & \\
\hline \hline
Computer Science & 8215 & 6987 & 104 & 80 & 50 \\
Electronic/Electric & 8690 & 7439 & 83 & 105 & 71 \\
Bioengineering & 6071 & 5552 & 81 & 75 & 50 \\
Mathematics & 6396 & 6100 & 105 & 61 & 41 \\
\hline 
\end{tabular}
\caption{\Acad\ dataset statistics (across the four chosen domains) including number of abstracts in the form of paper-\acad\ pairs, number of unique papers, number of unique \acad s, mean number of papers per \acad, and median number of papers per \acad.}
\label{table:stats}
\end{table*}

We conduct experiments using data in the \acad\ dataset for the four domains (Table~\ref{table:stats}) for training and testing, as reported in  Table \ref{table:trainstats} showing, for each of the four domains selected, the total number of labels (i.e. authors) and the number of training/testing unique papers. We also conduct experiments training on the \acad\ dataset restricted to the four domains (Table \ref{table:stats}) and testing on the grants in the four domains: Table~\ref{table:traingrantsstats} shows the number of grants awarded to \acad s from the selected departments who are found in the data from Table \ref{table:stats}. As for the abstracts,  (titles of) grants appear with multiple grant-holders, and thus each grant corresponds to a pair (referred to as \emph{grant pair} in Table~\ref{table:traingrantsstats}).

\begin{table}
\centering
\begin{tabular}{ cccc } 
\textbf{Domain} & \textbf{Grant pairs} & \textbf{Grants} & \textbf{\Acad s} \\ \hline \hline
Computer Science & 103 & 70 &  38\\
Electronic/Electric & 96 & 73 & 37 \\
Bioengineering & 111 & 95 & 38 \\
Mathematics & 99 & 90 & 46 \\
\hline 
\end{tabular}
\caption{Grant dataset statistics (across the four chosen domains) including number of grant pairs (in the form of grant-\acad\ pairs), number of unique grants, and number of unique \acad s.}
\label{table:traingrantsstats}
\end{table}

\begin{table}
\centering
\begin{tabular}{ ccccc } 
& \textbf{CS} & \textbf{EE} & \textbf{Bioeng} & \textbf{Maths} \\ \hline \hline
Labels (authors) & 104 & 83 & 81 & 105 \\
Training papers & 4,475 & 4,431 & 3,352 & 4,033 \\
Testing papers & 2,512 & 3,008 & 2,200 & 2,067 \\
\hline
\end{tabular}
\caption{Total number of labels and number of unique papers used for each round of training and  testing with the \acad\ dataset.}
\label{table:trainstats}
\end{table}

In our experiments in this section, data in the grant dataset is re-interpreted so that titles represent inputs and all the grant-holders the labels. Similarly, in the \acad\ dataset, data is re-interpreted so that the text of abstracts represent  inputs and all the authors (that are part of the department being analysed) represent the labels. We also ensure that each \acad\ has a minimum number of papers in the training set and a minimum number of papers in the testing set. In particular, an author with fewer than 25 publications will have 15 (randomly selected) abstracts used during training, an author with fewer than 50 publications will have 25 (randomly selected) abstracts used during training, an author with fewer than 100 publications will have 50 (randomly selected) abstracts used during training, and an author with more than 100 publications will have 100 (randomly selected) abstracts used during training. In all cases, the abstracts not selected for training are used for testing. At each training round with the \acad\ dataset, the split between training and testing data, resulting after having imposed the aforementioned restrictions, is as indicated in Table~\ref{table:trainstats}.

We  use ANN as baseline, considering unigrams and bigrams that appear in at least two abstracts and removing non-informative words such as pronouns and stop words. We represent features  as one-hot-encoding vectors where 1 indicates the presence of the unigram/bigram in text and 0 its absence. Our ANN has 3 Dense layers of 1024, 512, and 256 neurons, with ReLU as activation function. We use Dropout 0.3 after each of the first two dense layers. Our final layer is a dense layer with the number of classes given by the number of labels (\acad s) of each domain. In the final layer we use sigmoid and binary crossentropy, with Adam optimiser. We set the batch size to 32 and train for 50 epochs.

We compare the performances of the baseline ANN with pre-trained transformer-based \cite{Vaswani:17} models, given that they have had a remarkable impact in the NLP landscape and have proven to be highly effective in downstream tasks. We experiment with several such models to fine tune our \acad\ classification task:
\begin{itemize}
\item \textbf{BERT}: Bidirectional Encoder Representations from Transformers \cite{bert}; this model is trained through masked language modelling and next sentence prediction for a bidirectional learning of contextualized word embeddings;
\item \textbf{RoBERTa}: Robustly optimized BERT approach \cite{roberta} is a an optimized version of BERT where the next sentence prediction is removed and dynamic masking is used in the pre-training stage;
\item \textbf{ALBERT}: a lightweight version of BERT that has fewer parameters and higher training speed \cite{albert}; 
\item \textbf{DistilBERT}: is a lightweight version of BERT in which knowledge distillation is leveraged during the pre-training phase \cite{distilbert}; 
\item \textbf{SciBERT}: is a pre-trained language model based on BERT but trained on a large corpus of scientific text \cite{scibert}. The training set consists of full text of papers, not just abstracts, with 18\% of papers from the Computer Science domain and 82\% from the Biomedical domain.
\end{itemize}
For the experiments with transformer-based models, we set the sequence length to 200, batch size to 8, train for 10 epochs with early stopping and patience 3.

\begin{table*}
\centering
\begin{tabular}{ ccccccccc } 

& \multicolumn{2}{c}{\textbf{Computer Science}} &
\multicolumn{2}{c}{\textbf{Electronic/Electric}} & \multicolumn{2}{c}{\textbf{Bioengineering}} & \multicolumn{2}{c}{\textbf{Mathematics}}  \\ 

& \textbf{HL} & \textbf{Match} & \textbf{HL} & \textbf{Match} & \textbf{HL} & \textbf{Match} & \textbf{HL} & \textbf{Match} \\ \hline \hline

ANN & .009 & 73.8$\pm$0.018 & .009 & 77.1$\pm$0.025 & .008 & 77.8$\pm$0.010 & .006 & 70.8$\pm$0.020 \\
BERT & .050 & 81.8$\pm$0.012 & .062 & 89.7$\pm$0.008 & .067 & 87.9$\pm$0.006 & .053 & 66.0$\pm$0.030 \\
RoBERTa & .042 & 84.7$\pm$0.005 & .054 & 89.2$\pm$0.006 & .045 & 90.2$\pm$0.003 & .051 & 71.5$\pm$0.018 \\
ALBERT & .042 & 81.5$\pm$0.007 & .056 & 89.7$\pm$0.006 & .051 & 89.0$\pm$0.006 & .044 & 72.5$\pm$0.013 \\
DistilBERT & .025 & 85.0$\pm$0.003 & .026 & 87.6$\pm$0.006 & .024 & 87.8$\pm$0.006 & .027 & 83.0$\pm$0.009 \\
SciBERT & .029 & \textbf{88.3}$\pm$0.006 & .029 & \textbf{91.3}$\pm$0.005 & .025 & \textbf{92.4}$\pm$0.003 & .030 & \textbf{85.6}$\pm$0.010 \\
\hline
\end{tabular}
\caption{Average results of 5 runs  with abstracts in training data randomly selected. \textit{HL} stands for Hamming Loss. \textit{Match} stands for the average of the percentages of the number of authors correctly identified (mean and standard deviation).}
\label{table:nn}
\end{table*}

We perform five experiments for each classifier and we report the average results (see Table~\ref{table:nn}). In each experiment, we randomly select (as indicated earlier) the abstracts that are used as training data, with the remaining ones used as testing data. We run five experiments to ensure that our models can generalise. We report the Hamming loss (HL), the average of the percentages of  the number of authors correctly identified for each paper (Match)\footnote{For example,  the Match number for a publication with 3 \acad s of which only one is identified by our framework  is 1/3.} as well as the standard deviation to show that our models are stable and perform consistently on the task considered.  This resulted in 30 experiments for each of the tested domains, thus 120 experiments in total for the task of identifying authors of abstracts.
The results in Table \ref{table:nn} allow us to compare the performance of different multilabel neural network classifiers. For all domains, SciBERT outperforms the other transformer-based models as well as the ANN baseline. We believe this is due to the fact that SciBERT has been optimised on scientific text from the Computer Science and Biomedical domains.

\begin{table}
\centering
\begin{tabular}{ lcccc } 

\textbf{Method} & \textbf{CS} & \textbf{EE} & \textbf{Bioeng} & \textbf{Maths}  \\ 
\hline \hline

ANN & 52.5  & 49.8 & 64.3 & 39.4 \\
BERT & 56.7 & 61.2 & 64.3 & 39.6 \\
RoBERTa & 57.3 & 59.5 & 67.2 & 45.2 \\
ALBERT & 60.2 & 58.4 & 63.2 & 42.2 \\
DistilBERT & 55.3 & 58.4 & 63.9 & 45.3  \\
SciBERT & 74.2 & 74.7 & 78.7 & 52.2 \\
\hline
\textbf{B}aseline profile & 87.5 & 90.0 & 88.9 & 81.9 \\
\textbf{B} + \textbf{K}nowledge \textbf{b}ase  & 87.5 & 90.0 & 88.9 & 81.9 \\
\textbf{B} + \textbf{Emb}eddings & \textbf{92.4} & \textbf{90.4} & \textbf{89.9} & \textbf{83.0} \\
\textbf{B} + \textbf{Kb} + \textbf{Emb} & \textbf{92.4} & \textbf{90.4} & \textbf{89.9} & \textbf{83.0} \\
\hline
\end{tabular}
\caption{Results (i.e. averages of the percentages of the number of authors correctly identified) on predicting grant-holders from specific departments when using abstracts as training data for author prediction (top) and when using our framework (bottom). Here, our search method using just author profiles serves as the baseline (B).}
\label{table:nn-grants}
\end{table}

Finally, we compare the performances of the ANN and of the transformer-based models against our framework, to predict grant-holders, obtaining the results in Table \ref{table:nn-grants}. Here, all models are trained on the \acad\ dataset, and our framework is constructed from it, but all models and frameworks are tested on the grant dataset. Note that the grant-holders have been restricted to only those from the domains of analysis, as have \acad\ and abstracts from the \acad\ dataset.
This means that, for example, if an \acad\ from Computer Science and an \acad\ from Bioengineering are on the same grant, then, in our experiments, for that grant we would have only the Computer Science \acad\ when we test for the Computer Science domain, and similarly for the Bioengineering \acad.  Note that, here as well, SciBERT is the best performing model amongst the language-based models, but the performance is lower when tested on grants (see Table~\ref{table:nn-grants}) than when tested on abstracts (see Table~\ref{table:nn}). We believe this is due to the fact that these grant titles lack the information richness that can be found in abstracts. In particular,  grant titles have very few words, most of these words are umbrella terms or generic terms such as \emph{modular representation} as well as \emph{systems} and  \emph{methodology}. 

Given that our search framework empowered by domain-specific knowledge bases and word embeddings learnt from publication abstracts  places more focus on (all) individual words, it is better suited at finding  grant-holders, as shown in Table \ref{table:nn-grants}, where we see our framework using the academic profiles learnt from abstracts alone as baseline to which enhancements with knowledge base and word embeddings can be applied. Our framework outperforms the language-based models on all domains. When using embeddings, our framework obtains improvements that vary from 11.2\% for Bioengineering to 30.8\% for Mathematics. Adding knowledge base to the \acad s' profiles does not improve the results, which may be attributed to the fact that embeddings are already powerful enough, and in this case sufficient, at capturing relations existing in the data.

\subsection{Cross-domain identification of \acad s} \label{exp1}

Up until now, we have focused on authors from only four domains (out of 15).  This restriction  however limits the possibility of identifying cross-disciplinary research. In this section, we show that our (unsupervised) framework from Section \ref{search}, which identifies academic expertise given user queries, can achieve good performance on all grants, not limited to specific domains.

Table \ref{table:search} shows the percentage of predicted \acad s that are a subset of the \acad s associated to each grant (i.e. recall) and a performance breakdown of exact match for grants that have all ($G_{All}$, 2697 grants to test), one ($G_1$, 1895 grants to test), two ($G_2$, 579 grants to test), three ($G_3$, 144 grants to test), or more ($G_+$, 79 grants to test) \acad s as grant holders.
In the experiments, we report recall as academic search is defined as a recall-oriented task \cite{Kim:11}. In particular, for each grant, we run our framework, keep only the predicted \acad s that are also grant-holders, and then test whether we have an exact match. Note that, as in the experiments in the previous section, including embeddings improves the results obtained using a search against academic profiles alone.

\begin{table*}
\centering
\begin{tabular}{ cccccccc } 
\textbf{Kb} &  \textbf{Emb}  & \textbf{Recall} & $\mathbf{G_{All}}$ & $\mathbf{G_1}$ & $\mathbf{G_2}$ & $\mathbf{G_3}$ & $\mathbf{G_+}$ \\ \hline \hline
 \multicolumn{2}{c}{}  & 94.29 & 89.47 & 92.66 & 85.66 & 80.56 & 56.96 \\ 
\checkmark  &  & 94.29 & 89.47 & 92.66 & 85.66 & 80.56 & 56.96 \\ 
 & \checkmark & \textbf{94.88} & \textbf{90.25} & \textbf{93.30} & \textbf{86.36} & \textbf{81.25} & \textbf{62.03} \\ 
\checkmark & \checkmark & \textbf{94.88} & 90.03 & 93.25 & 86.01 & 80.56 & 59.49 \\ \hline
\end{tabular}
\caption{Results on predicting grant-holders using grant titles (with keywords concatenated, if any) as user queries. Here  our search method using just \acad\ profiles (first row) is the baseline. \textit{Recall} measures the percentage of predicted \acad s that are a subset of the \acad s associated to each grant, \textit{$G_{All}$} measures the percentage of grants for which our method identified all \acad s, \textit{$G_1$} measures the percentage of grants that have a single \acad\ which our method correctly identified. (Similarly \textit{$G_2$} focuses on grants with 2 \acad s, \textit{$G_3$} with 3 \acad s, and \textit{$G_+$} with more than 3 \acad s.)}
\label{table:search}
\end{table*}

\subsection{Properties} \label{sec:properties}

We define the following desirable properties for our search framework: (1) we aim to provide \textit{diverse} results, i.e. results that are not necessarily solely from the domain that one might expect or from the most senior \acad\ available in a domain; (2) we aim to provide \textit{novel} results, i.e., results that are not solely based on the terms that appear in the query but also from terms that are related to the query. Diversity is important as, depending on practical contraints, authors with the highest number of publications matching a query may have limited availability or belong to unsuitable departments (for example, if the query aims at identifying matches in specific departments). Novelty is important because it supports looking beyond simple keyword matches.

Below, we use $\A_\Q$ to stand for any of $R_\Q$ or $R_{\Q}^+$ (see Section~\ref{method}). \\
\textit{\textbf{Diversity}} can be formulated as follows:
$$\forall \Q \exists a_1,a_2 \in \A_\Q [ a_1\neq a_2 \wedge d(a_{1}) \neq d(a_{2}) \land p(a_{1}) \neq p(a_{2})].$$ 
Thus, diversity refers to the variety of identified \acad s with respect to the departments \OC{$d$} and positions \OC{$p$} held. A list of \acad s with low variety may not be of interest to the user. 
From the grants in our evaluation dataset with more than a single grant-holder, 35\% have \acad s from different departments and 51\% have \acad s holding different positions, e.g., professor vs lecturer/assistant professor vs researchers. Our framework is able to identify diverse departments and positions in 84\% of these cases. \\
\textit{\textbf{Novelty}} can be formulated as follows: $$\exists f [f \in V \land f \notin \QT \land f \in \mathcal{F}(\QT)]$$  where $f$ is a feature from an explanation vector $V$ and $\mathcal{F}(\QT)$ is a set of  features relevant to $\QT$, as specified by the source used to enhance the academic profiles, namely knowledge bases and/or embeddings. Our framework identifies not only \acad s based on matching features from $\QT$ with academic profiles but also other \acad s based on similarity of embeddings in the vector space. Thus, novelty referes to finding researchers, using knowledge bases and/or embeddings, who otherwise would not be discovered. For instance, ``deep learning for image segmentation" leads to authors of ``convolutional neural networks" because of the use of embeddings. Table \ref{table:search} shows that using embeddings leads to better results (i.e. more diverse and novel predictions). Because of these properties,  users of our system can use its predictions, along with relevance scores and explanations, to make  informative choices within and  across disciplines.

\section{Deployment}
\label{sec:new}

In this section we illustrate how our method can be deployed to obtain and explain answers for a variety of user queries (Section~\ref{illustrate}), as well as in combination with filtering mechanisms (Section~\ref{filter}), and discuss its computational efficiency (Section~\ref{efficient}).

\subsection{Explaining answers to queries}
\label{illustrate}

We analyse several queries we generated and illustrate explanations for the \acad s identified by our method, drawn from the explanation vectors and scores defined in Section~\ref{method}. These vectors use features from the query identified in the academic profiles as well as features obtained using the learnt embeddings, if applicable (we ignore the use of knowledge bases here as we have seen that they are not effective). The explanations result from checking the existence of  bigrams, followed by checking the existence of unigrams that form a bigram if that bigram is not found in the academic profile of the identified \acad s. Note that all example queries below are given as research areas/topics but could alternatively be structured as phrases.

Query: \textit{natural language processing}. Our method may return one \acad\ with the following explanation vector: \begin{quote}
\textit{language processing, natural language},
    \end{quote}
which gives an explanation score, $S_E$, of 20. Another \acad\ may, for example, have the explanation \textit{language, processing}, in which case $S_E$ is 2. If the explanation for one other \acad\ is solely \textit{language}, then $S_E$ is 1. The latter two explanations, \textit{language, processing} and  \textit{language}, may not indicate the best \acad s in terms of the expertise given the query, as these may be more related to \textit{programming languages and processing} rather than \textit{natural language processing}. However, our ranking, which takes into account $S_E$,  corrects this issue and forces the most relevant \acad s to be shown first (in our case the \acad\ who works on \emph{language processing, natural language}).

Query: \textit{biomedical image analysis}. Here, the maximum explanation score using academic profiles is 20, which can be achieved by \acad s who work on \textit{biomedical image, image analysis}. Further, there can be \acad s who work solely on \textit{image analysis}, with a $S_E$ of 10.
When using embeddings and searching the nearest 25 words for each word in the query in the academic profiles, we can obtain \acad s with the following explanation vector: \begin{quote}\textit{segmentation, neuroscience, landmark, image analysis, slice}.\end{quote}
Thus, the embeddings are useful in obtaining more information, by means of explanation vectors, as to which areas an \acad\ specialises in. These may lead to more relevant features and better results while also focusing the search and making it more specific.

Query: \textit{sentiment analysis speech language models}. An explanation vector based on analysing the academic profile may be: \begin{quote}\textit{language model, sentiment analysis, speech language, analysis, speech}
\end{quote}
with $S_E=32$ (3 bigrams and 2 unigrams), whereas an explanation that includes embeddings may be: \begin{quote}
\textit{textual, language, speech, model, visual, ensemble, video, analysis}.
\end{quote}

Query: \textit{verification on programming languages and program specification}. An explanation vector based on academic profiles may revolve around \textit{program specification, programming language, verification, language, program}. Two possible explanations constructed with the use of embeddings may be: 
\begin{quote}
\textit{toolchain, semantic, logic, predicate, specification, verification, symbolic}
    
\end{quote}
and 
    
\begin{quote}
\textit{satisfiability, execution, abstraction, optimisation, correctness refinement, verification, reasoning}.
   \end{quote}

Query: \textit{artificial intelligence}. Whilst this is an umbrella term for methods that enable machines to mimic human intelligence, also encompassing machine learning and deep learning, a search using this query may not produce adequate results. Using embeddings, the user can see topics related to \textit{artificial intelligence} and thus narrow and refine the search query. The following explanation may be given for an \acad: \begin{quote}
  \textit{bayes, deep learning, reinforcement learning, probability distribution, genetic algorithm, probabilistic model, autonomous, monte carlo}.  
\end{quote} This shows the importance of embeddings in identifying and explaining \acad s matching queries.

Note that our explanations can guide the users and help them understand the results, allowing them to filter out unwanted matches while also increasing their confidence in the system. In particular, from the explanation vector and the explanation score of a recommendation, users can better decide whether to go ahead and contact experts.

\subsection{Filtering}
\label{filter}

Different types of users may be looking for researchers at different points in their career. For example, students may need to identify suitable supervisors for projects, administrators may need to match funding opportunities with relevant researchers, and industry may be looking for opportunities for collaborative  work. Furthermore, researchers may need to identify collaborators from other disciplines than the ones they are in (i.e. they may want to exclude departments from the search).

To address these constraints, our framework can be deployed in combination with filtering capabilities, e.g. for department and position. Multiple departments and positions can be specified to:
\begin{itemize}
    \item select only those \acad s in the specified departments;
    \item select \acad s that are not in the specified departments;
    \item select only those \acad s that hold the specified positions;
    \item select \acad s that do not hold the specified positions.
\end{itemize}

In addition to filtering, users may also use the \acad\ and explanation score for browsing for an even better match for their interests.

\subsection{Efficiency}
\label{efficient}

In terms of space complexity, our framework needs to store the learnt embeddings which require memory space proportional to their size, $O(m \cdot d)$, where $m$ is the total number of features identified from \acad s' profiles and $d$ is the dimension of the vector (i.e. 100). Additionally, we store the computed profiles consisting of features that characterise each \acad, $O(n \cdot |\A|)$, as well as an index on the features towards the \acad s who have that feature in their profile for a faster retrieval.

Our framework is efficient in terms of adding publications of existing or new researchers. Whereas, when using neural models for multilabel classification, adding a new researcher leads to a modification of the existing list of labels which require the models to be retrained, 
in our framework,  a profile for a new \acad\ can be created and added to the already computed profiles. Indeed, our framework keeps track of the terms that represent the \acad s' profiles and the publication date of the papers in which these terms occur, making it simple and fast for new terms to be added to the existing profiles. As a result, the time complexity for insertion/addition is $O(1)$.
Furthermore, retrieving \acad s given a user query using profiles only can be done in $O(1)$ due to the feature indexing. Using word embeddings and/or knowledge base requires several calls with different search terms.

To summarise, our framework is computationally intensive when computing the researchers' profiles for the first time; adding new terms or researchers does not require our framework to re-compute profiles and can thus be done efficiently.

\section{Conclusion} \label{conclusion}

We proposed a framework for searching and exploring academic expertise. Our method is based on the creation of academic profiles by analysing the abstracts of their scientific publications, empowered by domain-specific knowledge bases and word embeddings learnt from the publications' abstracts. Our framework is able to return explanations for its recommendations, in the form of n-grams justifying matches. Moreover, it also works across-disciplines.

We envisage various avenues for future work. Firstly, we would like to conduct user studies to determine the usefulness of our framework, for instance in finding collaborators or PhD supervisors, and test the rating components of our algorithm. In particular, it would be helpful to assess whether knowledge bases, not useful for our evaluation in this paper, could instead prove useful with users.
Our framework also provides explanations for each \acad\ identified as related to the search query. As humans may make better judgments than machines when equipped with the right information, we plan to have a human-in-the-loop approach that requires human interaction to correct explanations in the case of false positives and thus improve the precision of our proposed method.
To support user experiments and human-in-the-loop developments, we plan to build an interactive user interface. Lastly, we foresee other sources of data that would be interesting to experiment with, including publications belonging to other universities and publications from selected departments from arXiv.\footnote{\url{https://arxiv.org/}}

\FT{\section*{Acknowledgements}
This research was carried out while the first author was at Imperial College London. The research was supported by Imperial College London's Faculty of Natural Sciences, Department of Computing and ICT.}

\bibliographystyle{named}
\bibliography{main}

\end{document}